\documentclass{article}

\usepackage{arxiv}
\usepackage{natbib}
\usepackage[utf8]{inputenc} 
\usepackage[T1]{fontenc}    
\usepackage{hyperref}       
\usepackage{url}            
\usepackage{booktabs}       
\usepackage{amsfonts}       
\usepackage{nicefrac}       
\usepackage{microtype}      
\usepackage{lipsum}		
\usepackage{graphicx}
\usepackage{doi}
\usepackage{color}
\usepackage{researchpack}
\usepackage[capitalise,nameinlink]{cleveref}
\usepackage{booktabs}
\usepackage{enumitem}
\usepackage{xspace}
\usepackage{multirow}
\usepackage{floatrow}
\usepackage[T1]{fontenc}
\usepackage{graphicx}
\usepackage[T1]{fontenc}
\usepackage{hyperref}
\usepackage{booktabs}
\usepackage{wrapfig}
\usepackage{amsmath, amssymb}
\usepackage[capitalise,nameinlink]{cleveref}
\usepackage{caption}
\usepackage{graphicx}
\usepackage{xcolor}
\usepackage[most]{tcolorbox}
\usepackage[misc]{ifsym}
\usepackage{multicol, multirow}
\usepackage{listings}
\usepackage{tikz}
\newtcolorbox{promptbox}{
	colback=blue!5!white,
	colframe=blue!60!black,
	title=Prompt,
	fonttitle=\bfseries,
	sharp corners,
	boxrule=0.8pt,
	left=6pt,
	right=6pt,
	top=6pt,
	bottom=6pt
}

\hypersetup{
	colorlinks=true,
	citecolor=blue,
	linkcolor=blue,
	filecolor=blue,
	urlcolor=blue,
}

\def\hb{\hbox to 11.5 cm{}}

\newcommand{\method}{\textsc{ConfGuide}\xspace}

\newcommand{\MimicCXR}{\texttt{Mimic-CXR-IV}\xspace}
\newcommand{\crcup}{\textsc{CRC++}\xspace}

	\title{Hybrid Decision Making via \\ Conformal VLM-generated Guidance} 
	
	\author{Debodeep Banerjee\\
		DI, University of Pisa\\
		DISI, University of Trento\\
		\And
		Burcu Sayin\\
		DISI, University of Trento\\
		\And
		Stefano Teso  \\
		CIMeC, university of Trento\\
		DISI, University of Trento\\
		\And
		Andrea Passerini \\
		DISI, University of Trento \\
	}
	
\begin{document}
		\maketitle
		\begin{abstract}
			Building on recent advances in AI, hybrid decision making (HDM) holds the promise of improving human decision quality and reducing cognitive load.
			We work in the context of \textit{learning to guide} (LtG), a recently proposed HDM framework in which the human is always responsible for the final decision: rather than suggesting decisions, in LtG the AI supplies (textual) \textit{guidance} useful for facilitating decision making.
			One limiting factor of existing approaches is that their guidance compounds information about all possible outcomes, and as a result it can be difficult to digest.
			%
			
			We address this issue by introducing \method, a novel LtG approach that generates more succinct and targeted guidance. To this end, it employs conformal risk control to select a set of outcomes, ensuring a cap on the \textit{false negative rate}. We demonstrate our approach on a real-world multi-label medical diagnosis task. Our empirical evaluation highlights the promise of \method. 
		\end{abstract}


	\section{Introduction}

	In high-stakes domains such as healthcare, law, and public safety, decision-making increasingly involves collaboration between human experts and AI systems. Fully automated decision-making is often inappropriate in these settings, as errors can be costly and many decisions require contextual judgment beyond statistical prediction. Hybrid decision-making (HDM) has therefore emerged as a promising paradigm for combining human and machine strengths. Yet, existing HDM approaches often fall short of delivering these benefits reliably in practice. In particular, many HDM techniques \citep{madras2018predict, mozannar2020consistent, keswani2022designing, verma2022calibrated, liu2022incorporating, wilder2021learning, de2020regression, raghu2019algorithmic, okati2021differentiable} assume a form of \textit{separation of responsibilities} between human and machine: for a given input, either the human or the AI system is assigned responsibility for the decision. As a result, these methods leave limited room for direct, joint collaboration on the same instance.

	To address this limitation, \citet{banerjee2024learning} introduced the \textit{learning to guide} (LTG) framework. Their method, \textsc{SLOG}, learns a guidance generator $\gamma(x)$ that maps an input $x$ to guidance $g$ intended to support human decision-making on that same input. In this setting, the AI system does not replace the human decision-maker, but instead contributes to the final decision through instance-specific guidance. However, \textsc{SLOG} requires fine-tuning of the underlying model, which can make deployment computationally expensive. Building on this line of work, \citet{banerjee2025medgellan} proposed \textsc{MedGellan}, a fine-tuning-free and purely inference-based framework, and demonstrated its effectiveness on medical ICD code prediction. Despite these advances, important challenges remain. In particular, while eliminating fine-tuning improves efficiency, the resulting guidance may still lack sufficiently clear argumentative structure, potentially leaving the end user uncertain about how to interpret or act on it.

	We present \method, an LLM-based pipeline for generating useful, well-structured guidance for multi-label decision-making tasks. Our approach first uses an ML model to produce pathology predictions and then applies conformal risk control (CRC) \citep{angelopoulos2022conformal} to select a prediction set of decision-relevant outcomes with a guaranteed (customizable) cap on the false negative rate.  Then it employs a VLM to generate arguments both in favor of and against each alternative decision. Because the generated guidance is grounded in risk-controlled prediction sets, \method provides a clearer argumentative basis for human judgment while reducing the risk of missing critical pathologies. We demonstrate the effectiveness of \method on multi-label pathology classification from chest X-ray images, where the system produces prediction-guided arguments that are ultimately presented to a physician for final decision-making. In this way, \method addresses both the problem of \textit{separation of responsibilities} and the safety concern of excessive false negatives.

	%
	\textbf{Contributions.}  Summarizing, we:
	\begin{itemize}
		\item Introduce \method, a finetuning-free pipeline for generating guidance for assisting human decision makers.
		\item Present a preliminary evaluation of \method on a challenging medical diagnosis task.
	\end{itemize}

	

	\section{Generating Conformal Guidance with \method}
	
	\paragraph{\textbf{Conformal risk control.}}  We first briefly review \emph{conformal risk control} (CRC) \citep{angelopoulos2022conformal}.  
	%
	CRC is a generalized version of classical conformal prediction \citep{papadopoulos2002inductive, vovk2005algorithmic, angelopoulos2023conformal} that provides a set of predictions by automatically calibrating a prediction threshold based on a user-specified \textit{risk}, such that the rate of missed findings does not overshoot a certain risk level at test time.
	We employ it to enforce a formal upper bound on the false negative rate (FNR) of the prediction set -- so as guarantee that \method won't overlook any likely present pathology.
	
	More generally, CRC works with any loss that is monotonically non-increasing with respect to a tunable conservativeness parameter, and has been shown to be effective in several settings, including multi-label classification.
	%
	Let $f_\theta(x) \in [0,1]^K$ denote a fixed pretrained model producing class-wise scores for input $x$. For a scalar parameter $\lambda \in [0,1]$, the prediction set is defined as: let \(C_\lambda(x) = \{y:f_y(x)>1-\lambda\}\) denote the prediction set produced for input \(x\) at threshold \(1-\lambda\), where \(\lambda\) controls the conservativeness of the prediction rule. Larger values of \(\lambda\) lower the threshold $1-\lambda$, yielding larger prediction sets. An optimal threshold \(1-\hat{\lambda}\) is then selected via CRC using a held-out calibration set.
	CRC requires the inference-time loss to be monotonically non-increasing in \(\lambda\).  Thankfully, our loss of choice --  the FNR -- does satisfy this requirement.
	For a predicted label set \(C_\lambda(x)\) and ground-truth label vector \(y \in \{0,1\}^K\), the FNR is defined as
	\[
	\mathrm{FNR}\big(C_\lambda(x), y\big)
	=
	\frac{\sum_{k=1}^K \mathbb{I}\{y_k = 1,\, k \notin C_\lambda(x)\}}
	{\max\left(1, \sum_{k=1}^K \mathbb{I}\{y_k = 1\}\right)}.
	\]

	
	
	\paragraph{\textbf{The role of $\lambda$ in CRC.}} Let $f_\theta(x)\in[0,1]^C$ denote a fixed trained model producing class probabilities (or scores).
	For a scalar parameter $\lambda\in[0,1]$, define the prediction rule
	\[
	\widehat{y}_c^{(\lambda)}(x)=\mathbf{1}\{ f_\theta(x)_c \ge 1-\lambda\}, \qquad c=1,\dots,C,
	\]
	i.e., $\lambda$ controls the decision threshold $t(\lambda)=1-\lambda$ (larger $\lambda$ yields a lower threshold and hence more predicted positives).
	Given a bounded, monotone loss $L(x,y;\lambda)\in[0,1]$ (e.g., per-example FNR) and an exchangeable calibration set
	$\{(X_i,Y_i)\}_{i=1}^n$, CRC selects
	\begin{equation}
		\label{crc_algo}
		\hat{\lambda}
		=\min\left\{\lambda\in\Lambda:\ \frac{\sum_{i=1}^n L(X_i,Y_i;\lambda)+1}{n+1}\le \alpha\right\}
	\end{equation}
	where $\alpha\in[0,1]$ is the target risk level and $\Lambda$ is a discrete grid of candidate $\lambda$ values. As our study centered around a problem related to clinical diagnosis, we considered the FNR to be our risk as missing a true pathology can turn out to be more fatal than falsely predicting an absent one.
	\begin{figure}
		\centering
		\includegraphics[width=0.75\linewidth]{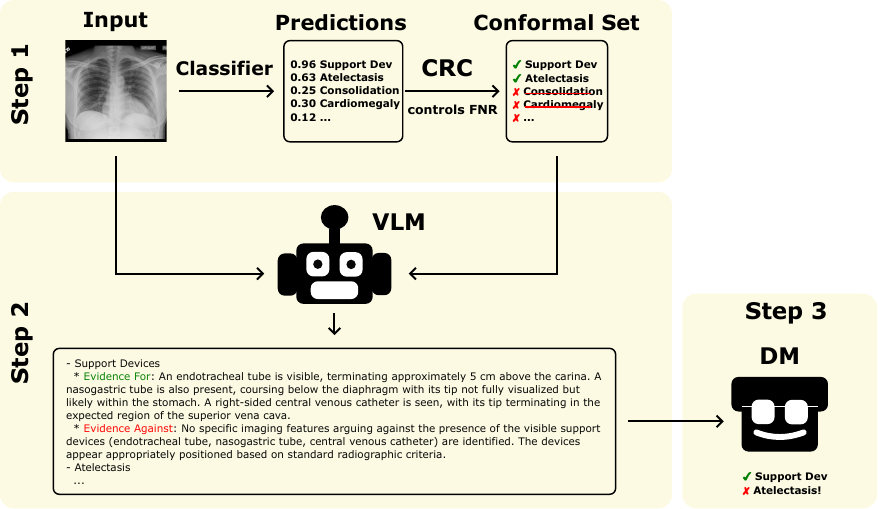}
		\caption{The functionality of \method is explained. The step 1 takes image as input. step ii takes image and the predictions as input and produces clinical guidance. In step iii, the doctor participates in evaluation with both image and the guidance.}
		\label{fig:placeholder}
	\end{figure}
	
	\paragraph{\textbf{\method.}}\method is designed to ensure that the decision maker takes the final decision while the underlying system provides guidance, maintaining a ceiling on the false negative rate as the target risk.
	%
	It consists of three interlinked elements: (i) a \textit{predictor}, (ii) an \textit{assistant}, and (iii) the \textit{decision maker} themselves. The first two elements are purely computational, while the the third one is the human in the loop. 
	These elements are invoked sequentially, cf. \Cref{fig:placeholder}.  In a first step, the predictor takes a chest X-ray as input and produces a probabilistic prediction for each candidate pathology.  Then the CRC algorithm (\Cref{crc_algo}) expands this set so that the FNR of the resulting \textit{conformal} prediction set is bounded at a prescribed level $\alpha$.
	In a second step, the assistant -- a state-of-the-art vision-language model -- receives the X-ray image and the conformal set of pathologies, and generates radiographic reasoning for and against the presence of each flagged pathology.
	These arguments are then presented to the human decision maker.  This design is motivated by the need to minimize anchoring bias --- the tendency of clinicians to over-rely on automated flags without critical evaluation.

	\section{Preliminary experiments}
	\label{sec:experiments}
	
	We empirically address the following research questions:
	\begin{itemize}
		\item[\textbf{Q1}] Does \method help in improving downstream decision quality? 
		\item[\textbf{Q2}] Does \method's guidance help compared to using the conformal set directly?
	\end{itemize}
	We implemented \method using Python and ran our experiments to generate the guidance on 2 NVIDIA A100 80 GiB GPUs and the simulated doctor models on an NVIDIA A100 40 GiB GPU.  We will publish our implementation and a template for the user study upon acceptance.

	
	
	\paragraph{\textbf{Dataset.}}We primarily used the \textsc{ChexPert} \citep{irvin2019chexpert} dataset for calibration and evaluation. For calibration, we utilize the official validation split of the dataset and reserved the test split for final evaluation. The calibration split is used to estimate $\hat{\lambda}$ and 
	subsequently select $\alpha$. Following \citet{tiu2022expert}, we discard lateral projections, retaining $202$ frontal images for calibration and $500$ for evaluation. The class distribution of the test set is reported in \Cref{tab:class_dist}.
	We leveraged the pretrained model provided by \citet{tiu2022expert}\footnote{\href{https://github.com/rajpurkarlab/CheXzero}{https://github.com/rajpurkarlab/CheXzero}}, trained using the \MimicCXR data set \citep{johnson2019mimic}, to predict the presence or absence of $14$ pathologies.

	\begin{table}[htbp]
		\centering
		\caption{Class distribution in the test set used this study.}
		\scalebox{0.75}{
			\label{tab:class_dist}
			\begin{tabular}{lcc}
				\toprule
				\textsc{Pathology} & \textsc{Positive} & \textsc{Negative} \\
				\midrule
				Atelectasis                & 153 & 347 \\
				Cardiomegaly               & 151 & 349 \\
				Consolidation              & 29  & 471 \\
				Edema                      & 78  & 422 \\
				Cardiomediastinum          & 253 & 247 \\
				Fracture                   & 5   & 495 \\
				Lung Lesion                & 8   & 492 \\
				Lung Opacity               & 264 & 236 \\
				No Finding                 & 62  & 438 \\
				Pleural Effusion           & 104 & 396 \\
				Pleural Other              & 4   & 496 \\
				Pneumonia                  & 11  & 489 \\
				Pneumothorax               & 9   & 491 \\
				Support Devices            & 261 & 239 \\
				\midrule
				Total                      & 1392 & 5608 \\
				\bottomrule
		\end{tabular}}
	\end{table}

	
	\paragraph{\textbf{Optimal Operating Point Selection.}}  The Conformal Risk Control (CRC) algorithm effectively reduces the False Negative Rate (FNR) by expanding the prediction set \(C_\lambda(x)\); however, this often comes at the cost of a high False Positive Rate (FPR). 
	To identify a significance level $\hat{\alpha}$ trading-off statistical safety and prediction efficiency, we define a selection procedure over a candidate set $\mathcal{A} = \{ \alpha_1, \alpha_2, \dots, \alpha_n \}$. For each $\alpha \in \mathcal{A}$, we first determine the calibrated threshold $\hat{\lambda}_{\alpha}$ using the Conformal Risk Control (CRC) framework using \Cref{crc_algo}. Let $\Bar{s_\alpha} = \frac{1}{n}\sum_{i=1}^{n}|C_\alpha(x_i)|$ be the average set size for 
	each $\alpha$ and $C_\lambda(x_i)$ be the cardinality of the prediction set for each input.
	
	Inspection of the empirical risk curve (see \Cref{fig:alpha-opt}) reveals four plateau regions: $\mathcal{P}_0 = [0.10, 0.15]$ with $\hat{R}=0.093$ and $\bar{s_\alpha}=11.57$, $\mathcal{P}_1 = [0.35, 0.40]$ with $\hat{R}=0.318$ and $\bar{s_\alpha}=8.5$, $\mathcal{P}_2 = [0.45, 0.55]$ with $\hat{R}=0.434$ and $\bar{s_\alpha}=7.3$, $\mathcal{P}_3 = [0.65, 0.70]$ with $\hat{R}=0.498$ and $\bar{s_\alpha}=4.5$, where $\hat{R}$ is the emprirical risk. Although $\mathcal{P}_0$ achieves the lowest empirical risk, we select  $\alpha^* = \min(\mathcal{P}_2) = 0.45$ as $\mathcal{P}_2$ is the longest plateau -- spanning three steps vs.\ two -- indicating greater stability of the risk  curve with respect to perturbations in $\alpha$. This robustness criterion reduces sensitivity to calibration noise, a desirable property in safety-critical clinical settings.
	
\begin{minipage}{0.48\textwidth}
	\centering
	\captionof{table}{Conformal Risk Control Performance at $\alpha = 0.45$}
	\label{tab:conformal_results}
	\scalebox{0.8}{
		\begin{tabular}{@{}lcc@{}}
			\toprule
			\textbf{Dataset} & \textbf{Target $\alpha$} & \textbf{Empirical FNR} \\ 
			\midrule
			Validation       & 0.45 & 0.4336 \\
			Test             & 0.45 & 0.4307 \\
			\bottomrule
	\end{tabular}}
\end{minipage}
\hfill
\begin{minipage}{0.48\textwidth}
	\centering
	\includegraphics[width=0.7\linewidth]{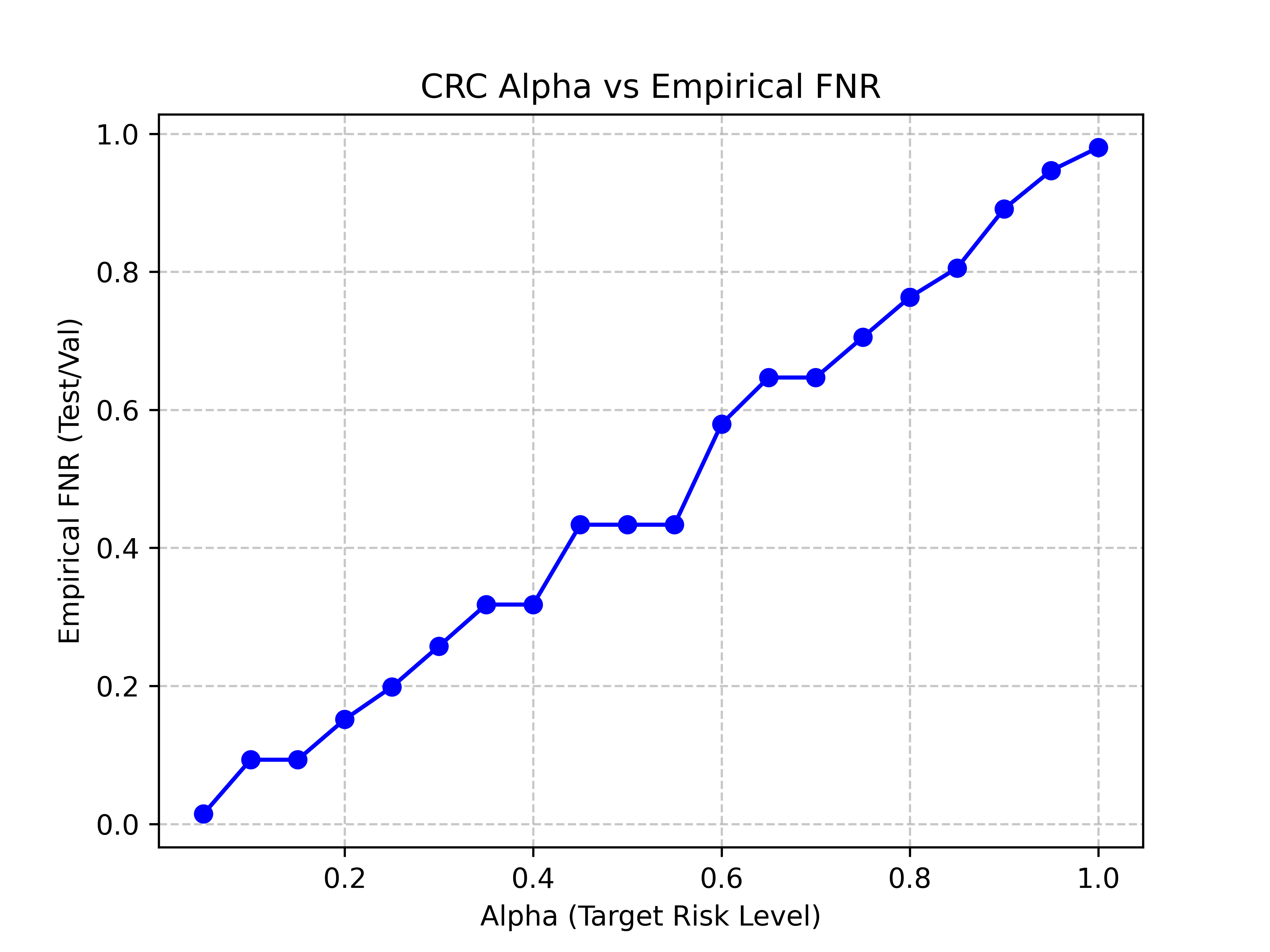}
	\captionof{figure}{Role of the $\alpha$ in determining the FNR}
	\label{fig:alpha-opt}
\end{minipage}
	
	\paragraph{\textbf{Guidance.}}  We retained the set of pathologies whose predictions satisfied the calibrated risk constraint at the selected threshold $\hat{\lambda}$. These CRC-filtered results were then provided as input to a SOTA  multimodal vision language model specifically designed to solve tasks belonging to medical deciplines. The VLM was asked to provide arguments \textit{in favor of} and \textit{against} the presence of each of the pathologies flagged with the previously mentioned classification model. In \Cref{tab:medllava_reasoning}, we present the nomenclature of the output of the task with an example. The prompt that is used to generate the guidance is shown in \Cref{box:guidance-prompt}.
	
	\paragraph{\textbf{Models.}}We performed the multi-label classification task using the pretrained models provided in \citet{tiu2022expert}, leveraging their zero-shot capabilities for identifying complex pathologies.. For the guidance generation, we use Google's MedGemma 27B \citep{sellergren2025medgemma} model. As a part of experiment, we simulate the doctor with two SOTA multi-modal language models, namely, GPI-4o-mini and Qwen-3-vl-8B \citep{bai2025qwen3}.
	
	\paragraph{\textbf{Metrics and evaluation strategy.}} The decisions received from the decision-maker VLM or the physician were compared with the ground truth labels and the qualities were assessed with the efficacy scores and confusion matrix. In one configuration, the full \method pipeline is utilized, i.e. the simulated doctor takes the guidance along with the image as input and the second configuration, the simulated doctor only receives the image as input. We define the second configuration as \crcup. Note that, both simulated decision maker VLM and physician do not have the opportunity to review the cases that were already marked as \textit{`absent'} with the CRC algorithm. Therefore, the decision-making parties can only review the cases that were flagged as \textit{`present'} with the CRC algorithm.

	\begin{tcolorbox}[fonttitle=\scriptsize, fontupper=\tiny, colback=yellow!10, colframe=purple!80!black, coltitle=white, label=box:guidance-prompt, title=Prompt for generating \textit{Guidance},left=1pt, right=1pt, top=1pt, bottom=1pt, boxsep=1pt ]
		{\tt {You are an expert radiologist reading a chest X-ray with a focus on balanced, evidence-based analysis.}
			
			{\tt Your task is to provide clinical reasoning BOTH for and against the presence of {item} based on visible radiographic features.}
				
				\tt CONTEXT:   1. {item} has been flagged as `present' by an automated system
					2. The X-ray is taken from AP view (anterior-posterior projection)
					3. Your reasoning will be reviewed by a clinician to make a final decision
					4. Your role is honest image interpretation, NOT confirming or defending the flag
					REQUIREMENTS:
					1. **Be specific** — Reference exact anatomical locations (e.g., ``right upper lobe", ``left costophrenic angle")
					2. **Use radiographic terminology** — opacity, consolidation, lucency, blunting, silhouette sign, air bronchogram, volume loss, tracheal deviation, pleural line
					3. **Report what you actually see** — Do NOT manufacture arguments for a side that isn't supported by the image
					4. **Asymmetric output is expected and correct** — if one side is weak, say so explicitly
					5. **Name mimics directly** — do not say "could be due to other causes", name the specific mimic
					6. **Connect features to diagnosis** — explain WHY a feature supports or contradicts {label\_name}
					7. **Be concise** — 2-3 sentences per side, most salient features only.
					
					\textbf{FOR ``reasons for presence''}:
					1. What specific features in this image support {label\_name}?
					2. Describe location, density, borders, radiographic signs
					3. If you see NO supporting features, state exactly: "No specific imaging features supporting {label\_name} are visible in this image"
					4. If evidence exists but is weak or atypical, explicitly say so
					FOR \textbf{"reasons against presence''}:
					5. Name the single most likely alternative explanation (mimic, normal variant, artifact)
					6. State which expected features of {label\_name} are absent
					7. Reference specific visible features that argue against it
					8. Do NOT use image quality or overlying lines/tubes as a generic excuse unless they genuinely and specifically obscure the relevant anatomical zone for {label\_name}
					
					CRITICAL RULES:
					1. Do NOT use the same observation to argue both sides
					2. Do NOT introduce findings unrelated to {label\_name}
					3. Do NOT speculate beyond what is directly visible
					4. Asymmetric output is correct — do not pad the weaker side to appear balance}
				
				Return ONLY valid JSON:
				{{"label": "{label\_name}", "favor": "...", "against": "..."}}
			\end{tcolorbox}
			\begin{table}[htbp]
				\centering
				\caption{An example of arguments in favour of and against the presence of selected pathologies for one single input.}
				\label{tab:medllava_reasoning}
				\scalebox{0.8}{
					\begin{tabular}{p{6em} p{6cm} p{6cm}}
						\toprule
						\textbf{Label} & \textbf{Reasons in Favor} & \textbf{Reasons Against} \\
						\midrule
						
						Atelectasis &
						There is increased opacity in the right lower lung field, particularly in the right cardiophrenic region, which could represent atelectasis. The right hemidiaphragm appears slightly elevated compared to the left, which can be associated with volume loss from atelectasis & The increased opacity in the right lower lung field is more suggestive of consolidation or pleural effusion rather than atelectasis, which typically presents with linear opacities or volume loss signs like crowding of vessels or fissures. There is no clear visualization of air bronchograms or significant tracheal deviation towards the right, which would be expected with substantial atelectasis. The opacity appears somewhat homogeneous and ill-defined, more consistent with fluid or inflammation. \\
						
						Cardiomegaly &
						The cardiac silhouette appears enlarged, particularly in the right heart border, which seems more prominent than expected. The overall width of the heart shadow at the level of the hemidiaphragm also appears increased. These findings suggest an enlarged cardiac size. &
						The AP projection inherently magnifies the heart size compared to a PA projection, potentially creating a false impression of cardiomegaly. The mediastinum appears widened, which could be due to the patient's rotation or increased mediastinal fat rather than true cardiac enlargement. The cardiac silhouette borders are not sharply defined, making precise assessment difficult. \\
						
						Edema &
						The image shows bilateral, diffuse, patchy opacities predominantly in the lower lung zones, with some perihilar distribution. These opacities have ill-defined borders and appear somewhat confluent, which can be consistent with pulmonary edema. There is also some blunting of the left costophrenic angle, potentially indicating a small pleural effusion, which can accompany edema.  &
						The opacities are somewhat asymmetric, appearing more pronounced on the right side. While edema can be asymmetric, the pattern is not classic for cardiogenic edema (which typically involves perihilar haze and Kerley B lines, neither of which are clearly visible here). The opacities could also represent multifocal pneumonia or aspiration, which are important differential diagnoses given the patchy nature and lack of clear upper lobe predominance often seen in edema. \\
						
						\bottomrule
				\end{tabular}}
			\end{table}
			
			\section{Results}
			\label{sec:results}
			\begin{table}[htbp]
				\centering
				\caption{Overall performance comparison across pipeline configurations.}
				\scalebox{0.8}{
					\label{tab:llm_results}
					\begin{tabular}{lccccccc}
						\toprule
						& &\multicolumn{3}{c}{\textsc{Macro}} & \multicolumn{3}{c}{\textsc{Micro}} \\
						\cmidrule(lr){3-5} \cmidrule(lr){6-8}
						\textsc{Config} & \textsc{Model} & \textsc{Pr} & \textsc{Rec} & \textsc{$F_1$} & \textsc{Pr} & \textsc{Rec} & \textsc{$F_1$} \\
						\midrule
						Standard ($\lambda$=0.5)  & -- & 37.48 & 66.28 & 33.59 & 25.36 & 60.20 & 35.68 \\
						CRC                       & -- & 36.58 & 71.87 & 34.48 & 24.94 & 65.01 & 36.05 \\
						\midrule
						\crcup      & GPT 4o-mini & 39.00 & 47.80 & 34.77 & 40.84 & 55.56 & 47.07 \\
						\method     & GPT 4o-mini & 39.00 & 54.77 & \textbf{35.48} & 44.30 & 59.41 & \textbf{50.76} \\
						\midrule
						\crcup      & Qwen3-vl-8B & 37.31 & 59.96 & 34.97 & 31.99 & 62.57 & 42.34 \\
						\method     & Qwen3-vl-8B & 38.32 & 54.02 & \textbf{35.22} & 41.06 & 60.56 & \textbf{48.94} \\
						\bottomrule
				\end{tabular}}
			\end{table}
			
			\begin{table}[htbp]
				\centering
				\caption{Per-pathology Precision, Recall, and F1 scores for CRC vs \method.}
				\scalebox{0.75}{
					\label{tab:per_pathology}
					\begin{tabular}{lrrrrrr|rrrrrr}
						\toprule
						& \multicolumn{3}{c}{\textsc{CRC}} & \multicolumn{3}{c|}{\textsc{\method~(GPT-4o-mini)}} & \multicolumn{3}{c}{\textsc{CRC}} & \multicolumn{3}{c}{\textsc{\method~(Qwen)}} \\
						\cmidrule(lr){2-4} \cmidrule(lr){5-7} \cmidrule(lr){8-10} \cmidrule(lr){11-13}
						\textsc{Pathology} & \textsc{Pr} & \textsc{Rec} & \textsc{$F_1$} & \textsc{Pr} & \textsc{Rec} & \textsc{$F_1$} & \textsc{Pr} & \textsc{Rec} & \textsc{$F_1$} & \textsc{Pr} & \textsc{Rec} & \textsc{$F_1$} \\
						\midrule
						Atelectasis                & 54.04 & 83.01 & 65.46 & 58.65 & 79.74 & \textbf{67.59} & 54.04 & 83.01 & 65.46 & 58.65 & 79.74 & \textbf{67.59} \\
						Cardiomegaly               & 76.35 & 74.83 & \textbf{75.59} & 76.35 & 74.83 & \textbf{75.59} & 76.35 & 74.83 & \textbf{75.59} & 76.35 & 74.83 & \textbf{75.59} \\
						Consolidation              & 10.73 & 96.55 & 19.31 & 13.85 & 93.10 & \textbf{24.11} & 10.73 & 96.55 & 19.31 & 12.39 & 93.10 & \textbf{21.86} \\
						Edema                      & 45.65 & 80.77 & \textbf{58.33} & 48.44 & 39.74 & 43.66 & 45.65 & 80.77 & \textbf{58.33} & 45.83 & 56.41 & 50.57 \\
						Enlarged Cardiomediastinum & 92.62 & 54.55 & \textbf{68.66} & 92.62 & 54.55 & \textbf{68.66} & 92.62 & 54.55 & \textbf{68.66} & 92.62 & 54.55 & \textbf{68.66} \\
						Fracture                   & 0.99  & 80.00 & 1.96  & 5.56  & 40.00 & \textbf{9.76}  & 0.99  & 80.00 & 1.96  & 5.88  & 20.00 & \textbf{9.09}  \\
						Lung Lesion                & 1.93  & 100.00 & 3.79 & 6.74  & 75.00 & \textbf{12.37} & 1.93  & 100.00 & 3.79 & 4.23  & 75.00 & \textbf{8.00}  \\
						Lung Opacity               & 73.04 & 95.45 & 82.76 & 83.15 & 85.98 & \textbf{84.54} & 73.04 & 95.45 & 82.76 & 82.86 & 87.88 & \textbf{85.29} \\
						No Finding                 & 5.75  & 40.32 & 10.06 & 10.09 & 37.10 & \textbf{15.86} & 5.75  & 40.32 & 10.06 & 8.04  & 37.10 & \textbf{13.22} \\
						Pleural Effusion           & 48.24 & 92.31 & 63.37 & 49.23 & 92.31 & \textbf{64.21} & 48.24 & 92.31 & 63.37 & 48.98 & 92.31 & \textbf{64.00} \\
						Pleural Other              & 0.46  & 25.00 & \textbf{0.91} & 0.00  & 0.00  & 0.00  & 0.46  & 25.00 & \textbf{0.91} & 0.00  & 0.00  & 0.00  \\
						Pneumonia                  & 2.65  & 81.82 & 5.13  & 4.21  & 81.82 & \textbf{8.00}  & 2.65  & 81.82 & 5.13  & 3.65  & 72.73 & \textbf{6.96}  \\
						Pneumothorax               & 2.57  & 88.89 & \textbf{5.00} & 0.00  & 0.00  & 0.00  & 2.57  & 88.89 & \textbf{5.00} & 0.00  & 0.00  & 0.00  \\
						Support Devices            & 97.06 & 12.64 & 22.37 & 97.06 & 12.64 & 22.37 & 97.06 & 12.64 & 22.37 & 97.06 & 12.64 & 22.37 \\
						\bottomrule
				\end{tabular}}
			\end{table}

			\begin{table}[htbp]
				\centering
				\caption{Per-pathology Precision, Recall, and F1 scores for \method~$-$~guidance vs \method.}
				\scalebox{0.75}{
					\label{tab:ablation_pathology}
					\begin{tabular}{lrrrrrr|rrrrrr}
						\toprule
						& \multicolumn{6}{c|}{\textsc{GPT-4o-mini}} & \multicolumn{6}{c}{\textsc{Qwen3-VL-8B}} \\
						\cmidrule(lr){2-7} \cmidrule(lr){8-13}
						& \multicolumn{3}{c}{\textsc{\crcup}} & \multicolumn{3}{c|}{\textsc{\method}} & \multicolumn{3}{c}{\textsc{\crcup}} & \multicolumn{3}{c}{\textsc{\method}} \\
						\cmidrule(lr){2-4} \cmidrule(lr){5-7} \cmidrule(lr){8-10} \cmidrule(lr){11-13}
						\textsc{Pathology} & \textsc{Pr} & \textsc{Rec} & \textsc{$F_1$} & \textsc{Pr} & \textsc{Rec} & \textsc{$F_1$} & \textsc{Pr} & \textsc{Rec} & \textsc{$F_1$} & \textsc{Pr} & \textsc{Rec} & \textsc{$F_1$} \\
						\midrule
						Atelectasis                & 56.52 & 59.48 & 57.96 & 58.65 & 79.74 & \textbf{67.59} & 53.98 & 79.74 & 64.38 & 58.65 & 79.74 & \textbf{67.59} \\
						Cardiomegaly               & 77.69 & 62.25 & 69.12 & 76.35 & 74.83 & \textbf{75.59} & 76.35 & 74.83 & \textbf{75.59} & 76.35 & 74.83 & \textbf{75.59} \\
						Consolidation              & 12.75 & 89.66 & 22.32 & 13.85 & 93.10 & \textbf{24.11} & 12.00 & 93.10 & 21.26 & 12.39 & 93.10 & \textbf{21.86} \\
						Edema                      & 49.22 & 80.77 & \textbf{61.17} & 48.44 & 39.74 & 43.66 & 45.99 & 80.77 & \textbf{58.60} & 45.83 & 56.41 & 50.57 \\
						Enlarged Cardiomediastinum & 92.59 & 49.41 & 64.43 & 92.62 & 54.55 & \textbf{68.66} & 92.62 & 54.55 & \textbf{68.66} & 92.62 & 54.55 & \textbf{68.66} \\
						Fracture                   & 11.11 & 20.00 & \textbf{14.29} & 5.56  & 40.00 & 9.76  & 0.00  & 0.00  & 0.00  & 5.88  & 20.00 & \textbf{9.09}  \\
						Lung Lesion                & 1.02  & 25.00 & 1.96  & 6.74  & 75.00 & \textbf{12.37} & 2.30  & 100.00 & 4.49 & 4.23  & 75.00 & \textbf{8.00}  \\
						Lung Opacity               & 78.14 & 82.58 & 80.29 & 83.15 & 85.98 & \textbf{84.54} & 73.57 & 92.80 & 82.08 & 82.86 & 87.88 & \textbf{85.29} \\
						No Finding                 & 18.09 & 27.42 & \textbf{21.79} & 10.09 & 37.10 & 15.86 & 5.75  & 40.32 & 10.06 & 8.04  & 37.10 & \textbf{13.22} \\
						Pleural Effusion           & 48.54 & 79.81 & 60.36 & 49.23 & 92.31 & \textbf{64.21} & 50.30 & 81.73 & 62.27 & 48.98 & 92.31 & \textbf{64.00} \\
						Pleural Other              & 0.00  & 0.00  & 0.00  & 0.00  & 0.00  & 0.00  & 0.46  & 25.00 & \textbf{0.91} & 0.00  & 0.00  & 0.00  \\
						Pneumonia                  & 3.32  & 63.64 & 6.31  & 4.21  & 81.82 & \textbf{8.00}  & 3.36  & 81.82 & 6.45  & 3.65  & 72.73 & \textbf{6.96}  \\
						Pneumothorax               & 0.00  & 0.00  & 0.00  & 0.00  & 0.00  & 0.00  & 8.70  & 22.22 & \textbf{12.50} & 0.00  & 0.00  & 0.00  \\
						Support Devices            & 97.06 & 12.64 & 22.37 & 97.06 & 12.64 & 22.37 & 97.06 & 12.64 & 22.37 & 97.06 & 12.64 & 22.37 \\
						\bottomrule
				\end{tabular}}
			\end{table}

			\subsection{RQ1. \method helps in improving quality of downstream decision making. }
			
			Empirical results in \Cref{tab:llm_results} demonstrate that \method substantially improves downstream decision quality over both the standard threshold and CRC alone. \method with GPT-4o-mini achieves a micro-$F_1$ of 50.76, representing improvements of $14.71\%$ and $15.08\%$ over CRC and standard thresholding respectively. Consistent results with Qwen ($+12.90\%$ over CRC) suggest that gains are attributable to guidance quality rather than model-specific reasoning capacity.
			
			
			The modest macro-$F_1$ gain ($+1\%$) relative to micro-$F_1$ ($+14.71\%$) is explained by class imbalance (\Cref{tab:class_dist}): frequent pathologies dominate 
			micro-$F_1$ and consistently benefit from guidance, while equal class weighting in macro-$F_1$ limits the overall gain. Per-pathology results for GPT in \Cref{tab:per_pathology} confirm improvements in 10 out of 14 pathologies, with notable gains in Fracture ($+7.80$), Lung Lesion ($+8.58$), and Lung Opacity ($+1.78$). Degradation is observed in rare pathologies such as Pneumothorax and Pleural Other, where limited prevalence and subtle radiographic appearance make precise guidance generation challenging.
			
			\subsection{RQ2. \method helps in improving quality of downstream decision making against when the doctor is provided only the CRC sets without Guidance.}
			To isolate the contribution of the generated guidance, we compare \method against an ablated variant in which the decision model receives only the image without 
			guidance (\crcup). As shown in \Cref{tab:llm_results}, \method consistently outperforms \crcup under both GPT-4o-mini and Qwen, 
			confirming that the guidance --- rather than the filtering model alone --- drives the observed improvements. Per-pathology analysis in \Cref{tab:ablation_pathology} 
			shows that \method with GPT-4o-mini outperforms its corresponding \crcup in 8 out of 14 classes, ties in 3, and underperforms in 3, with degradation concentrated 
			in rare pathologies where guidance generation is inherently challenging. The \method with Qwen3, outperforms its counterpart in 10 classes. However,\method's consistent under-performance in two classes namely `Pleural other' and `Pneumothorax' need to be investigated further.

			\section{Discussion}
			
			\paragraph{Simulation.} In the \Cref{sec:results}, simulation studies demonstrates that \method can boost the quality of downstream decision making task. By combining  conformal risk control with VLM driven differential reasoning, the pipeline achieves significant gain in recovering false positives, while maintaining majority of the true positives provided with the CRC algorithm. The discrepancy between the micro and macro scores broadly stems from the data imbalance and the pipeline's shortcomings identify those classes, especially the `Pneumothorax' and `Pleural other'.   
			
			\paragraph{\textbf{User study.}} While the current study is primarily focused on investigation under a simulated set-up, a preliminary user study -- involving two clinicians with $15$ decision each -- highlighted the promising of \method. Specifically, \method helped the doctors in correctly identifying a subset of pathologies, for a minority of classes it failed to provide required assistance. We acknowledge however that a large-scale human study is necessary to draw reliable conclusions. This is a critical next step in our study.
			
			\section{Conclusion and future work}
			\label{sec:conclusion}
			
			In this study, we introduce a novel hybrid decision making approach, \method, that employs conformal risk control and prompt-based vision language model to generate structured guidance useful for down-stream human decisions.  Our experiments indicate that \method can successfully improve decisions in a simulated scenario.  In future research, we plan to conduct a larger scale user study and also evaluate improvements in agreement between multiple doctors due to guidance.


			\bibliography{reference}

@article{angelopoulos2022conformal,
  title={Conformal risk control},
  author={Angelopoulos, Anastasios N and Bates, Stephen and Fisch, Adam and Lei, Lihua and Schuster, Tal},
  journal={arXiv preprint arXiv:2208.02814},
  year={2022}
}

@article{tiu2022expert,
  title={Expert-level detection of pathologies from unannotated chest X-ray images via self-supervised learning},
  author={Tiu, Ekin and Talius, Ellie and Patel, Pujan and Langlotz, Curtis P and Ng, Andrew Y and Rajpurkar, Pranav},
  journal={Nature biomedical engineering},
  volume={6},
  number={12},
  pages={1399--1406},
  year={2022},
  publisher={Nature Publishing Group UK London}
}

@article{johnson2019mimic,
  title={{MIMIC-CXR-JPG-chest Radiographs with Structured Labels (version 2.0.0)}},
  author={Johnson, Alistair and others},
  journal={PhysioNet},
  year={2019}
}

@book{vovk2005algorithmic,
  title={Algorithmic learning in a random world},
  author={Vovk, Vladimir and Gammerman, Alexander and Shafer, Glenn},
  year={2005},
  publisher={Springer}
}

@inproceedings{papadopoulos2002inductive,
  title={Inductive confidence machines for regression},
  author={Papadopoulos, Harris and Proedrou, Kostas and Vovk, Volodya and Gammerman, Alex},
  booktitle={European conference on machine learning},
  pages={345--356},
  year={2002},
  organization={Springer}
}

@article{raghu2019algorithmic,
  title={The algorithmic automation problem: Prediction, triage, and human effort},
  author={Raghu, Maithra and Blumer, Katy and Corrado, Greg and Kleinberg, Jon and Obermeyer, Ziad and Mullainathan, Sendhil},
  journal={arXiv:1903.12220},
  year={2019}
}

@article{madras2018predict,
  title={{Predict Responsibly: Improving Fairness and Accuracy by Learning to Defer}},
  author={Madras, David and others},
  journal={NeurIPS},
  year={2018}
}

@inproceedings{mozannar2020consistent,
  title={Consistent estimators for learning to defer to an expert},
  author={Mozannar, Hussein and Sontag, David},
  booktitle={ICML},
  year={2020}
}

@inproceedings{de2020regression,
  title={Regression under human assistance},
  author={De, Abir and Koley, Paramita and Ganguly, Niloy and Gomez-Rodriguez, Manuel},
  booktitle={AAAI},
  year={2020}
}

@inproceedings{wilder2021learning,
  title={Learning to complement humans},
  author={Wilder, Bryan and others},
  booktitle={IJCAI},
  year={2021}
}

@article{liu2022incorporating,
  title={Incorporating uncertainty in learning to defer algorithms for safe computer-aided diagnosis},
  author={Liu, Jessie and others},
  journal={Scientific Reports},
  year={2022}
}

@inproceedings{verma2022calibrated,
  title={Calibrated learning to defer with one-vs-all classifiers},
  author={Verma, Rajeev and Nalisnick, Eric},
  booktitle={ICML},
  year={2022}
}

@article{keswani2022designing,
  title={Designing closed human-in-the-loop deferral pipelines},
  author={Keswani, Vijay and others},
  journal={arXiv:2202.04718},
  year={2022}
}

@article{okati2021differentiable,
  title={Differentiable learning under triage},
  author={Okati, Nastaran and others},
  journal={NeurIPS},
  year={2021}
}

@article{banerjee2024learning,
  author={Banerjee, Debodeep and Teso, Stefano and Sayin, Burcu and Passerini, Andrea},
  title={Learning to guide human decision makers with vision-language models},
  journal={arXiv preprint arXiv:2403.16501},
  year={2024}
}

@article{sellergren2025medgemma,
  title={Medgemma technical report},
  author={Sellergren, Andrew and Kazemzadeh, Sahar and Jaroensri, Tiam and Kiraly, Atilla and Traverse, Madeleine and Kohlberger, Timo and Xu, Shawn and Jamil, Fayaz and Hughes, C{\'\i}an and Lau, Charles and others},
  journal={arXiv preprint arXiv:2507.05201},
  year={2025}
}

@inproceedings{irvin2019chexpert,
  title={Chexpert: A large chest radiograph dataset with uncertainty labels and expert comparison},
  author={Irvin, Jeremy and Rajpurkar, Pranav and Ko, Michael and Yu, Yifan and Ciurea-Ilcus, Silviana and Chute, Chris and Marklund, Henrik and Haghgoo, Behzad and Ball, Robyn and Shpanskaya, Katie and others},
  booktitle={Proceedings of the AAAI conference on artificial intelligence},
  volume={33},
  number={01},
  pages={590--597},
  year={2019}
}

@article{banerjee2025medgellan,
  title={MedGellan: LLM-Generated Medical Guidance to Support Physicians},
  author={Banerjee, Debodeep and Sayin, Burcu and Teso, Stefano and Passerini, Andrea},
  journal={arXiv preprint arXiv:2507.04431},
  year={2025}
}

@article{angelopoulos2023conformal,
  title={Conformal prediction: A gentle introduction},
  author={Angelopoulos, Anastasios N and Bates, Stephen},
  journal={Foundations and Trends in Machine Learning},
  volume={16},
  number={4},
  pages={494--591},
  year={2023},
  publisher={Emerald Publishing Limited}
}

@article{bai2025qwen3,
  title={Qwen3-vl technical report},
  author={Bai, Shuai and Cai, Yuxuan and Chen, Ruizhe and Chen, Keqin and Chen, Xionghui and Cheng, Zesen and Deng, Lianghao and Ding, Wei and Gao, Chang and Ge, Chunjiang and others},
  journal={arXiv preprint arXiv:2511.21631},
  year={2025}
}
			\bibliographystyle{plainnat}

			\end{document}